\documentclass[letterpaper, 10 pt, conference]{ieeeconf}  
\usepackage{graphicx}
\usepackage{placeins}

\usepackage{xurl}
\usepackage{ragged2e}
\usepackage{amssymb}
\usepackage{amsmath}
\usepackage{multirow}
\usepackage{url}
\usepackage{array}
\usepackage{subcaption}
\usepackage[normalem]{ulem}

\IEEEoverridecommandlockouts                              

\title{\LARGE \bfseries
Learning Cooperative Bimanual Dexterous Grasp of Large Objects with Tactile Sensing 
}

\author{Ziming Li$^{1}$, Mingxuan Wu$^{2}$, Jiaqi Zhang$^{1}$, Hongfei Li$^{3}$,
Yan Gan$^{2}$, Deqiang Ouyang$^{2}$, and Ning Wang$^{2,*}$%
\thanks{This work was supported by the National Natural Science Foundation of China
under Grants 62572080 and 62101079, the Natural Science Foundation of Chongqing
(No. CSTB2023NSCQ-MSX1020), and the Sichuan Science and Technology Program
(No. 2025YFHZ0084).}%
\thanks{$^{1}$Ziming Li and Jiaqi Zhang are with the School of Computer Science,
The University of Auckland, Auckland 1010, New Zealand
(e-mail: {\small\texttt{lzm07072024@163.com}};
{\small\texttt{zjiaqi0717@gmail.com}}).}%
\thanks{$^{2}$Mingxuan Wu, Yan Gan, Deqiang Ouyang, and Ning Wang are with the
College of Computer Science, Chongqing University, Chongqing 401331, China
(e-mail: {\small\texttt{wmx@stu.cqu.edu.cn}};
{\small\texttt{shiyangancq@cqu.edu.cn}};
{\small\texttt{deqiangouyang@cqu.edu.cn}};
{\small\texttt{nwang5@cqu.edu.cn}}).}%
\thanks{$^{3}$Hongfei Li is with the College of Electronic and Information Engineering,
Southwest University, Chongqing 400715, China
(e-mail: {\small\texttt{hongfli@126.com}}).}%
\thanks{*Corresponding author: {\small\texttt{nwang5@cqu.edu.cn}}}%
}

\begin{document}

\maketitle
\thispagestyle{empty}
\pagestyle{empty}

\begin{abstract}


Bimanual dexterous grasping of large objects is a critical challenge in robotic manipulation. 
However, most existing studies focus on sequential manipulation rather than cooperative grasping, and methods addressing such bimanual tasks have largely been limited to simulation. 
These limitations stem from the difficulty of acquiring full 3D object models and generating physically plausible grasping actions. 
To fill this gap, we propose a real-world bimanual grasping framework that includes: a multimodal dataset capturing joint angles, visual observations and force signals; a Denoising Diffusion Probabilistic Model (DDPM)-based module that generates joint-level grasp configurations from segmented point clouds; and an execution strategy that integrates motion planning with online grasp refinement to ensure physical stability and feasibility.
Our approach enables the synthesis of executable bimanual grasps from single-view inputs, reducing dependence on complete 3D object models and ensuring stable real-world performance.
Experiments on a dual-arm robot demonstrate high success rates across unseen objects with varying geometries and poses, and ablation studies confirm the contributions of key components of our system.
Codes and dataset can be found at \url{https://github.com/zhangdana483/real_bi_dex_grasp/}.

\end{abstract}

\section{INTRODUCTION}

Embodied intelligence emphasizes the tight coupling between perception, action, and interaction within the physical world.
In this context, grasping has become a core component of robotic manipulation, serving as a fundamental skill that enables intelligent agents to act purposefully and autonomously. While most robotic systems have achieved reliable grasping with parallel-jaw \cite{DBLP:conf/icra/ONeillRMGPLPGMJ24_1} or suction grippers \cite{DBLP:journals/ral/OBrienKKL24_2}, such end-effectors are limited in handling complex, large-scale, or irregular objects. Dexterous hands, with high degrees of freedom and tactile sensitivity, have emerged as powerful tools to tackle fine-grained manipulation tasks, enabling in-hand adjustment \cite{DBLP:conf/icra/YuanCQHYL0L024_3}, soft-actuated skills \cite{DBLP:journals/trob/ZhouHDAL25_4}, and contact-rich interactions \cite{lakshmipathy2024contactmpc_5}.

\begin{figure}[t]
   \centering
   \includegraphics[width=3.3in]{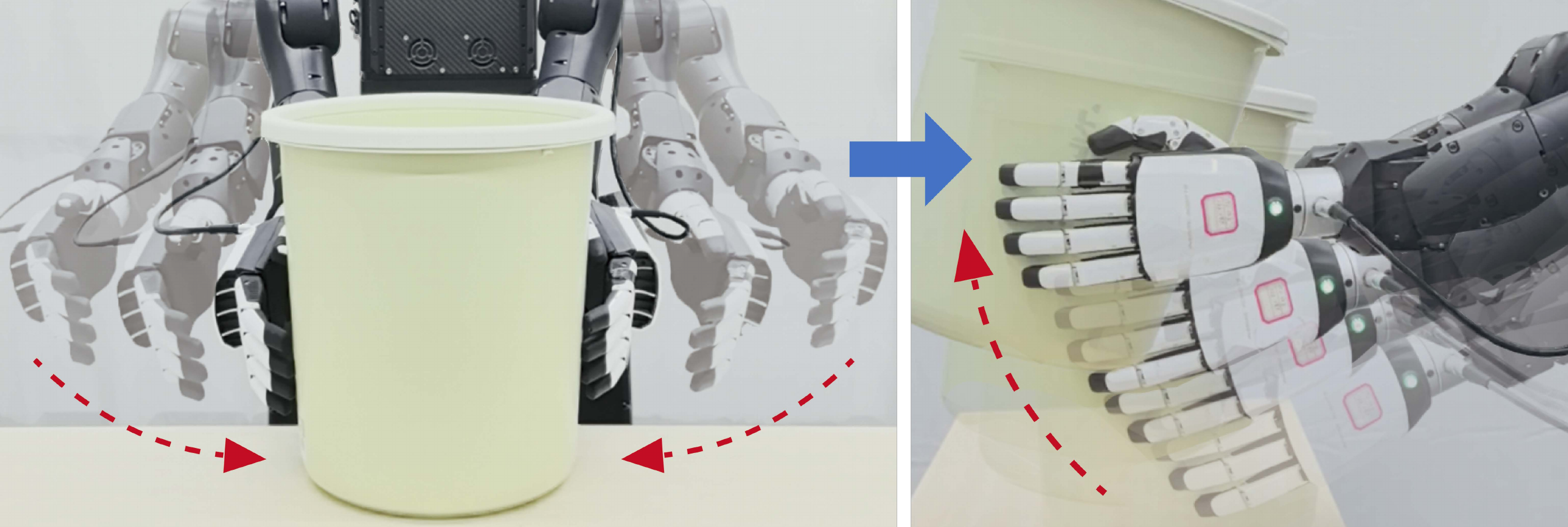}
   \caption{Many large daily objects, such as bins and balls, are too wide to be grasped by one hand. This requires coordinated bimanual strategies that allow a robot to enclose the object from both sides and lift it securely.}
   \label{fig:introduction}
\end{figure}

Despite this progress, most dexterous grasping research focuses on unimanual settings, assuming the object can be enclosed by a single hand \cite{DBLP:journals/thms/Feix0SDK16_6}. In contrast, humans frequently engage in bimanual grasping to lift, hold, and stabilize large or heavy objects, such as boxes, buckets, or sports balls, by enclosing them with both arms, as illustrated in Figure~\ref{fig:introduction}. 
This bimanual capability is essential for handling objects that are too wide, heavy, or structurally simple (e.g., lacking handles) for one-handed manipulation. 
Extending robotic dexterous grasping to such bimanual scenarios represents a critical step toward more general-purpose embodied agents.

Most existing studies on bimanual dexterous manipulation focus on sequential, task-oriented scenarios such as pouring or folding, which are fundamentally different from the problem of jointly lifting a single large object with both hands. 
Although a few works have explored this setting, such as BimanGrasp-DDPM \cite{DBLP:journals/ral/ShaoX24_7}, which generates stable grasp poses using stochastic optimization and diffusion models, existing methods remain predominantly constrained to idealized simulation environments and face critical limitations in real-world applications. A key limitation lies in their reliance on complete 3D mesh models of target objects as input. In practice, especially for large-scale objects, robots often operate with only partial observations from a single-view depth camera, where severe self-occlusion and restricted field of view make full shape reconstruction difficult or even impossible. While multi-view reconstruction techniques have been proposed \cite{DBLP:conf/cvpr/WangCKV0N25_8}, they typically require the robot to physically move around the object to capture multiple views, which is often impractical in real-world settings due to time constraints and limited accessibility.

Moreover, most existing bimanual dexterous grasping approaches generate grasp poses without considering whether the robot can physically reach or execute the configuration. For instance, some synthesized grasps involve supporting the object from underneath, which is infeasible when the object rests on a tabletop. 
Another critical shortcoming is the lack of execution-time adaptability in the grasping process. Relying solely on visual input, current systems are unable to detect contact misalignment or to subsequently compensate for small geometric deviations during execution, often leading to insufficient contact engagement and unstable grasps in real-world settings.
Taken together, these limitations significantly restrict the applicability of current bimanual grasp synthesis methods outside of simulation environments.

To overcome these limitations, we present a real-world bimanual grasp synthesis and execution framework for large objects using dexterous hands. Our system is built on a dual-arm humanoid robot equipped with dexterous hands and a head-mounted RGB-D camera. We collect a real-world bimanual grasping dataset via teleoperation using Apple vision Pro, covering various large objects (e.g., boxes, buckets, balls). Each grasp sequence includes visual observations, joint states, force sensing, and successful lift-and-place trials. Based on that, we train a diffusion model to predict the target joint configuration of arms and hands from single-view segmented point clouds, avoiding reliance on full object meshes. At execution time, the robot performs motion planning to reach the target state, followed by online grasp adjustment, allowing for contact refinement and grasp stabilization. Our contributions are summarized as follows:

\begin{itemize}

\item We propose a real-world bimanual grasping framework with a DDPM-based model for grasp synthesis and an execution strategy combining motion planning and online grasp refinement.

\item We establish a dataset of bimanual grasping through teleoperation, with multimodal data including joint angles, vision observation and force sensing.

\item We validate our framework on a real-world dual-arm robot, showing high success rates on unseen objects with diverse poses.

\end{itemize}

\section{Related Work}

\subsection{Dexterous Grasp}

Dexterous grasping enables robotic systems to manipulate a wide range of objects with precision and flexibility. Early work relied on analytical models based on geometric and physical constraints \cite{DBLP:journals/ral/LiuLJZZ22_10, DBLP:books/daglib/0073732_11, DBLP:conf/icra/PonceSBM93_12, DBLP:conf/icra/RosalesSGB12_14}, but suffered from scalability issues due to high-dimensional optimization and limited robustness in practice. Data-driven approaches have since become the dominant paradigm. Regression-based methods \cite{DBLP:conf/rss/LiuP0GM20_15, DBLP:conf/cvpr/XuWZ0024_16} predict grasp poses directly from object input, but often lack diversity and fail to generalize to novel shapes. Generative models \cite{DBLP:conf/iccv/JiangLW021_17, DBLP:conf/cvpr/XuWZLSSWGWCLYW23_18} offer richer grasp distributions, and recent diffusion-based approaches \cite{DBLP:conf/cvpr/HuangWLJLZLZ23_19, DBLP:conf/eccv/LuKLLYHH24_20, DBLP:journals/corr/abs-2407-09899_21} have shown strong capabilities in synthesizing realistic, diverse poses. Besides, DexGraspAnything \cite{DBLP:conf/cvpr/ZhongJYM25_22} integrates physics awareness into diffusion models, improving grasp robustness and execution viability.

Learning-based performance heavily depends on data. Simulated datasets, generated via tools like GraspIt! \cite{DBLP:journals/ram/MillerA04_23} and IsaacGym \cite{DBLP:conf/nips/MakoviychukWGLS21_24} using eigengrasp \cite{DBLP:conf/cvpr/HassonVTKBLS19_25, DBLP:conf/rss/LiuP0GM20_15, DBLP:journals/ral/LundellVK21_26} or optimization-based sampling \cite{DBLP:conf/rss/LiuP0GM20_15, DBLP:conf/cvpr/XuWZLSSWGWCLYW23_18, DBLP:conf/icra/LiLLGZYH23_27, DBLP:conf/icra/WangZCXLLW23_28}, are scalable but often limited in diversity and deployability. In contrast, teleoperated real-world datasets provide more natural and grounded grasps, though they are harder to scale. Recent works \cite{DBLP:conf/corl/ShawBP22_30, DBLP:conf/rss/SivakumarSP22_31} map human hand motions to robots, offering a more efficient way to collect large-scale data.

While these methods advance dexterous grasp synthesis, they largely focus on unimanual settings. Extending to coordinated bimanual grasping introduces new challenges in inter-arm control and reachability. BimanGrasp-DDPM \cite{DBLP:journals/ral/ShaoX24_7} explores bimanual grasp generation in simulation using diffusion models, but it is limited to simulated settings without physical validation. In contrast, our work is conducted on real robots with a multimodal dataset, enabling feasible bimanual grasping in the real world.

\begin{figure*}[!htbp]
   \centering
   \includegraphics[width=1.0\textwidth]{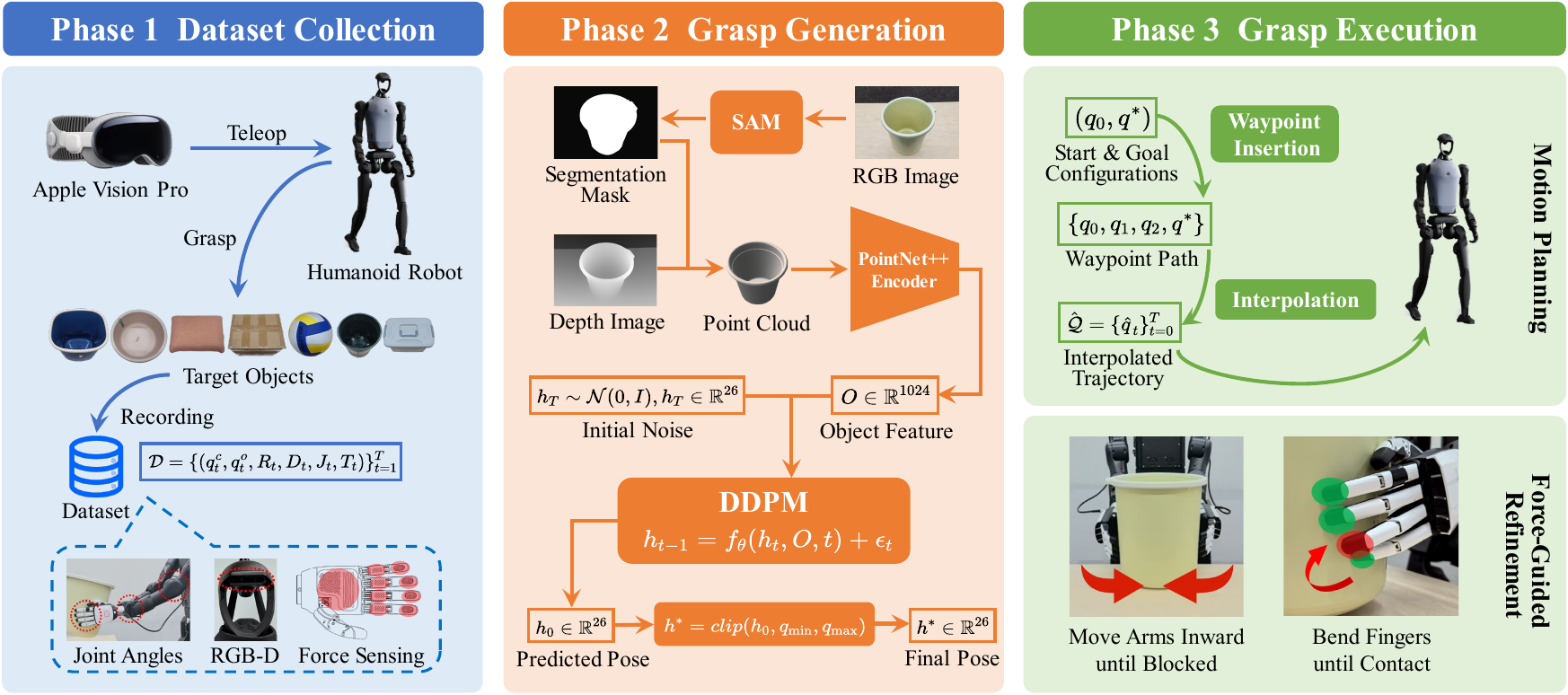}
   \caption{Pipeline of our method. The framework consists of three phases. In Phase 1, a teleoperated humanoid robot performs real-world bimanual grasps on large objects, and the multimodal data is continuously recorded, including visual observations, joint angles, and force sensing, which are then aggregated into a structured dataset. In Phase 2, a diffusion model is trained on this dataset to predict target joint configurations from segmented point clouds. These predictions are then passed to Phase 3 for execution via motion planning and force-guided refinement, enabling stable grasping in physical environments.}
   \label{fig:framework}
\end{figure*}

\subsection{Bimanual Manipulation}

Bimanual manipulation enables coordinated use of both arms, commonly studied in two forms: collaborative actions (e.g., folding, pouring) where hands play complementary roles; and cooperative grasping, where both arms hold a single object. These differ in control demands and have been explored with both grippers and dexterous hands.

Recent dual-arm gripper methods include COMBO-Grasp \cite{DBLP:journals/corr/abs-2502-08054_32} for constraint-guided occluded grasping, PerAct2 \cite{DBLP:journals/corr/abs-2407-00278_34} for transformer-based bimanual tasks, YOTO \cite{DBLP:journals/corr/abs-2501-14208_35} for one-shot diffusion skills, Grasplargeflat \cite{DBLP:conf/icra/WangK25} for RL-based grasping of large flat objects, and Bi-Touch \cite{DBLP:journals/ral/LinCYLLZL23} for tactile bilateral grasping. These methods rely on simple rigid grippers, which differ fundamentally from dexterous hands.

Dexterous bimanual manipulation has increasingly focused on sequential, task-oriented control. Datasets like OakInk2 \cite{DBLP:conf/cvpr/00010ZMXL0L24_36} and Text2HOI \cite{DBLP:conf/cvpr/ChaKYB24_37} capture fine-grained dual-hand interactions. Models such as AsymDex \cite{DBLP:journals/corr/abs-2411-13020_39} and DexMachina \cite{DBLP:journals/corr/abs-2505-24853_40} adopt role asymmetry or curriculum learning to mimic human-like coordination.

Despite progress in sequential tasks, fewer works address bimanual grasping of a single large object, which requires stable, opposing-hand enclosure. ViSiL-HD \cite{DBLP:journals/corr/abs-2502-20396} achieves dexterous real-world dual-arm manipulation through reinforcement learning, including the task of bimanual coordinated grasping of a single object, but its capability is confined to specific shapes like cuboids and lacks generalization. BimanGrasp-DDPM \cite{DBLP:journals/ral/ShaoX24_7} explores this task via diffusion-based synthesis, but relies on full object meshes, lacks reachability planning, and omits online adjustment, limiting its real-world applicability. 

In contrast, our approach leverages a real-world multimodal dataset and achieves executable bimanual dexterous grasps from partial single-view observations while incorporating force feedback for generalizable and stable real-world performance, thereby enabling more reliable grasp of large objects in real-world settings.

\section{METHOD}

We provide an overview of our method in Figure~\ref{fig:framework}. Given a single-view RGB-D observation of a large object, along with the segmented mask obtained from a visual perception model, our goal is to generate and execute a physically feasible bimanual grasp using two dexterous hands. To achieve this, we propose a three-phase framework: collecting a real-world multimodal grasping dataset via teleoperation, training a diffusion model to predict target joint configurations from segmented point clouds, and executing the predicted grasp with motion planning and force-based refinement.

\subsection{Dataset Collection}

To support learning-based bimanual dexterous grasp synthesis, we construct a real-world dataset through teleoperated demonstrations using a dual-arm robotic system. We use the Unitree H1-2 humanoid robot, with 7 Degrees of Freedom (DOF) in each arm, and two Inspire RH56DFTP dexterous hands, each providing 6 DOF. The hands are equipped with multimodal sensing capabilities, including joint torque sensors on each finger and distributed tactile sensors along the palmar surfaces of all finger links and the palm.

Teleoperation is implemented with Apple Vision Pro, which captures wrist poses and hand keypoints via sensor-fusion tracking. After coordinate transformation from OpenXR to the robot URDF convention, hand keypoints are retargeted to dexterous hand joints, and wrist poses are solved by inverse kinematics for dual-arm joint angles. The system is provided by Unitree, with real-time aligned vision from a head-mounted RealSense D435i.

Each bimanual manipulation trial consists of a continuous sequence of actions, including: moving the hands to the target object, closing fingers to establish a grasp, lifting the object, and finally placing it down. During this process, multimodal data is recorded at 15 Hz, comprising the commanded joint angles $q_t^c$, the actual joint angles $q_t^o$, RGB images $R_t$, depth images $D_t$, joint torques $J_t$, and tactile signals $T_t$. The dataset is formally represented as:

\begin{equation}
\mathcal{D}=\{(q_t^c,q_t^o,R_t,D_t,J_t,T_t)\}_{t=1}^T.
\end{equation}

\begin{figure}[t]
   \centering
   \includegraphics[width=3.2in]{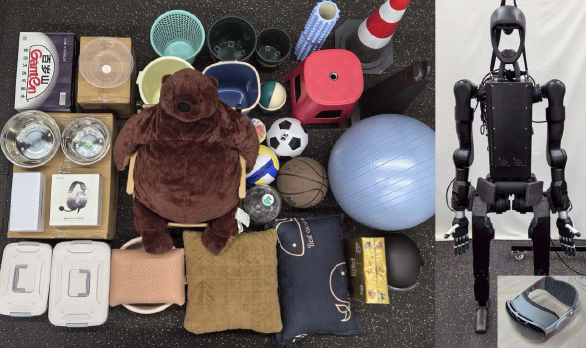}
   \caption{Hardware platform and all the grasped objects.}
   \label{fig:hardwareANDobjects}
\end{figure}

We collect grasp demonstrations on 33 different large objects, each with multiple pose variations (ranging from 6 to 20) to enhance diversity and generalization, resulting in a total of 353 grasps. Figure~\ref{fig:hardwareANDobjects} shows the hardware platform and all the grasped objects.

In addition, for the purpose of training our grasp generation model, we perform two post-processing steps on the time series data of each grasping sequence: the RGB image from the first frame is processed using ‌Segment Anything Model (SAM) to obtain the object mask, and a specific frame is manually annotated as the moment when grasping is just completed but the object has not yet been lifted. These two frames serve as inputs to the training pipeline. The same segmentation method is also applied during inference.


\subsection{Grasp Generation}

We predict bimanual grasp poses from RGB-D input by segmenting the object, extracting a partial point cloud, encoding its shape, and using a diffusion model to generate joint angles with joint limit clipping.

To extract object-level geometric features from real-world visual input, we define a preprocessing function $f_\mathrm{encoder}$ that takes as input an RGB image $I$, a depth map $D$, and the camera intrinsic matrix $K$, and outputs a global object feature vector $\mathcal{O}\in\mathbb{R}^{1024}$:

\begin{equation}
\mathcal{O}=f_\mathrm{encoder}(I,D,K).
\end{equation}

The RGB image $I\in\mathbb{R}^{H\times W\times3}$ is first processed using the Segment Anything Model (SAM) \cite{DBLP:conf/iccv/KirillovMRMRGXW23_41} to obtain a binary mask $\mathcal{M}\in\{0,1\}^{H\times W}$, which identifies the pixels belonging to the target object. These pixels are then extracted from the corresponding depth image $D\in\mathbb{R}^{H\times W}$, and are back-projected into 3D space using the camera intrinsic matrix $K\in\mathbb{R}^{3\times3}$. For each foreground pixel $(u,v)$ satisfying $\mathcal{M}(u,v)=1$, the corresponding 3D point $\mathbf{p}\in\mathbb{R}^3$ is computed as:

\begin{equation}
\mathbf{p}=\mathcal{M}(u,v)\cdot D(u,v)\cdot K^{-1}[u,v,1]^T.
\end{equation}


Here, $D(u,v)$ denotes the depth value at pixel location $(u,v)$, and the binary mask $\mathcal{M}(u,v)$ ensures that only object pixels contribute to the point cloud. The resulting set of points forms a partial object point cloud $\mathcal{P}=\{\mathbf{p}_i\}_{i=1}^N\subset\mathbb{R}^3$, where $N$ is the number of foreground pixels. This point cloud is then passed through a PointNet++ encoder \cite{DBLP:conf/nips/QiYSG17} to obtain the global geometric representation $\mathcal{O}\in\mathbb{R}^{1024}$, which serves as the condition for subsequent grasp pose generation.

Given the encoded object representation $\mathcal{O}\in\mathbb{R}^{1024}$, we aim to predict a 26-dimensional grasp pose vector $\mathbf{h}^{0}\in\mathbb{R}^{26}$, which includes the joint angles of two 7-DoF robotic arms and two 6-DoF dexterous hands. To predict bimanual grasp poses conditioned on object geometry, we adopt a Denoising Diffusion Probabilistic Model (DDPM) \cite{DBLP:conf/nips/HoJA20_43}, which transforms Gaussian noise into a 26-dimensional joint configuration through iterative denoising. Unlike direct regression, diffusion models iteratively denoise predictions conditioned on object features and noise levels, decomposing the task into simpler steps while progressively modeling the output structure.

The diffusion process gradually corrupts the ground-truth pose $\mathbf{h}^0$ into a Gaussian noise vector $\mathbf{h}^{T}$ over a fixed number of time steps $T$, using a forward noising process defined as:

\begin{equation}
q(\mathbf{h}^t|\mathbf{h}^0)=\mathcal{N}\left(\mathbf{h}^t;\sqrt{\bar{\alpha}_t}\mathbf{h}^0,(1-\bar{\alpha}_t)\mathbf{I}\right),\quad t=1,\ldots,T.
\end{equation}

Here, $\mathbf{h}^{t}\in\mathbb{R}^{26}$ denotes the noised pose at step $t$, and $\bar{\alpha}_t=\prod_{i=1}^t\alpha_i$ is the product of predefined noise scaling coefficients $\alpha_{i}\in(0,1)$. The model learns a reverse denoising process that progressively reconstructs $\mathbf{h}^0$ from $\mathbf{h}^{T}$, conditioned on the object feature $\mathcal{O}$. Specifically, at each denoising step $t$, a neural network $\epsilon_{\theta}$ predicts the noise component $\epsilon$ added to the original pose:

\begin{equation}
\epsilon_\theta:(\mathbf{h}^t,t,\mathcal{O})\mapsto\hat{\epsilon}.
\end{equation}

The model is trained to minimize the mean squared error between the predicted and actual noise, using the objective:

\begin{equation}
L_\theta=\mathbb{E}_{t,\mathbf{h}^0,\epsilon}\left[\left\|\epsilon-\epsilon_\theta\left(\sqrt{\bar{\alpha}_t}\mathbf{h}^0+\sqrt{1-\bar{\alpha}_t}\epsilon,t,\mathcal{O}\right)\right\|^2\right]\text{,}
\end{equation}
where $\epsilon\sim\mathcal{N}(0,\mathbf{I})$ is sampled Gaussian noise. After training, the model generates grasp poses by sampling an initial noise vector $\mathbf{h}^{T}\sim\mathcal{N}(0,\mathbf{I})$ and iteratively applying the learned reverse process to recover a clean pose estimate $\mathbf{h}^0$.

To ensure that the predicted joint angles lie within the physical range of each actuator, we apply joint limit clipping to the output. Given the lower and upper bounds $\mathbf{h}_{\min},\mathbf{h}_{\max}\in\mathbb{R}^{26}$, we compute the final executable grasp pose $\mathbf{h}^*=\mathrm{clip}\left(\mathbf{h}^0,\mathbf{h}_{\min},\mathbf{h}_{\max}\right)$. This ensures that each joint angle in $\mathbf{h}^*$ satisfies the hardware constraints.

\subsection{Grasp Execution}

After getting the predicted grasp configuration from the generation stage, we execute the grasp in two steps: an initial motion planning phase that brings arms and hands into the target pose, followed by a force-guided refinement phase that ensures stable and comprehensive contact with the object.

Given the predicted full-configuration pose $h^*\in\mathbb{R}^{26}$, we extract the target joint angles of the dual arms as $q^{*}\in\mathbb{R}^{14}$. Let $q_{0}\in\mathbb{R}^{14}$ denote the initial configuration with both arms in their natural resting pose. To generate a feasible trajectory from $q_{0}$ to $q^{*}$, we first construct an intermediate waypoint sequence $\mathcal{Q}=\{q_0,q_1,q_2,q^*\}$, where $q_1$ lifts the arms laterally to avoid potential collisions with the tabletop, and $q_2$ is defined by translating $q^*$ backward along the palm’s normal vector to provide a pre-grasp offset, ensuring that both hands can slowly approach the object from opposite sides along the grasping force direction, avoiding unintended contact that might push or misalign the object during motion. Linear interpolation is then applied between each consecutive pair in $\mathcal{Q}$ to construct a dense trajectory:

\begin{equation}
\hat{\mathcal{Q}}=\{\hat{q}_t\}_{t=0}^T,\quad\hat{q}_t\in\mathbb{R}^{14}\text{,}
\end{equation}
which is streamed to the robot controller at 50 Hz. We assign a non-uniform velocity profile across different segments of the trajectory, where faster speeds are used for the initial lifting and approach $(q_0\to q_1\to q_2)$ and slower motion is applied during the final insertion $(q_{2}\to q^{*})$. Once the target arm pose is reached, the remaining 12 joint angles from $h^*$ are used to actuate the finger joints for grasp execution.

Once the arms reach the target pose and complete the initial grasp, we perform a force-guided refinement to enhance contact stability and robustness. The first stage involves closing the arms until rigid contact is achieved. Let $p_{L},p_{R}\in\mathbb{R}^{3}$ be the positions of the left and right palms. We compute the midpoint $p_m=\frac{1}{2}(p_L+p_R)$ and iteratively move each palm toward $p_m$ along the vectors $p_m-p_{L}$ and $p_m-p_{R}$, with low-speed control until the relative motion between iterations falls below a small threshold, indicating that the arms are physically blocked by the object.

In the second stage, we refine individual finger poses based on distributed tactile feedback and joint torque sensing. For each finger $i\in\{1,\ldots,10\}$, let $\theta_i^{(t)}$ denote the joint angle at iteration $t$, and define the contact state as:

\begin{equation}
\mathrm{Contact}(i)=(J_i>\tau_{\mathrm{th}})\vee\left(\max_jT_{i,j}>\delta_{\mathrm{th}}\right)\text{,}
\end{equation}
where $J_{i}$ is the torque reading at the base joint of the $i$-th finger, and $T_{i,j}$ is the tactile reading from the $j$-th sensor on that finger. $\tau_{\mathrm{th}}$ and $\delta_{\mathrm{th}}$ are predefined thresholds for torque and tactile signal strength, respectively.

We denote the set of fingers that have not yet made contact at time step $t$ as $\mathcal{I}_{\mathrm{free}}^{(t)}=\{i\mid\mathrm{Contact}(i)=\mathrm{False}\}$. The joint angles of these fingers are incrementally updated by a small step $\Delta\theta_{i}$, yielding the update rule:

\begin{equation}
\theta_i^{(t+1)}=\theta_i^{(t)}+\Delta\theta_i,\quad\forall i\in\mathcal{I}_\mathrm{free}^{(t)}.
\end{equation}

As the five fingers bend, the thumb simultaneously moves inward. This iterative process continues until all fingers reach stable contact, enabling the robot to achieve reliable and enriched surface engagement across the entire grasp.

It is worth emphasizing that, although force feedback is additionally used during execution, this does not contradict the “single-view” claim in the title, because grasp synthesis relies solely on one head-mounted RGB-D observation, consistent with practical robotic deployment. Besides, although our setting assumes relatively simple scenes, the framework can be extended to more complex real-world environments using open-vocabulary object detection and collision-aware motion planning.


\section{EXPERIMENT}

\subsection{Experimental Setup}

We evaluate our grasp synthesis and execution pipeline on a real-world dual-arm robotic platform under structured but unconstrained tabletop settings. All experiments are performed using physical robots, with the same hardware setup as used during dataset collection. Our aim is to test whether grasps generated from a single RGB-D image are physically executable and robust across different object shapes and poses.

Each evaluation trial begins with a single RGB-D frame from a fixed Realsense D435i depth camera, using SAM to perform instance segmentation, followed by a dedicated selection step to retain the segmented region corresponding to the target object. Since the tabletop background is uniformly colored and objects are generally positioned in or around the central region of the field of view, we discard the largest segment located at the image border and merge all small or medium segments located near the image center. This merged region is treated as the grasp target and is back-projected using the aligned depth map to form a partial point cloud.

\begin{table}[h]
\centering
\caption{Parameters during each phase.} 
\label{tab:parameters}
\renewcommand{\arraystretch}{1.2}
\begin{tabular}{ccc}
\hline
\textbf{Category}               & \textbf{Parameter}         & \textbf{Value} \\ \hline
\multirow{8}{*}{Training/Inference} & Diffusion Steps            & 200             \\
                                & Epochs        & 50000             \\
                                & Batch Size                 & 16             \\
                                & Optimizer                  & AdamW             \\
                                & Learning Rate              & $2\times10^{-4}$             \\
                                & Hidden Dimension                 & 512             \\
                                & Condition Dimension            & 1024             \\ 
                                & Inference Steps                 & 200             \\ \hline
\multirow{5}{*}{Execution}      & Command Frequency          & $50\,\mathrm{Hz}$             \\
                                & Arm Velocity (Fast Phase)  & $0.5\,\mathrm{m/s}$             \\
                                & Arm Velocity (Slow Phase)  & $0.15\,\mathrm{m/s}$             \\
                                & Finger Angular Speed               & $1.57\,\mathrm{rad/s}$             \\
                                & Force Threshold (Contact) & $2.9\,\mathrm{N}$             \\ \hline
\end{tabular}
\end{table}

Each segmented point cloud is downsampled and encoded using the PointNet++. The denoising model is a 4-layer MLP with residual-free layers and SiLU-based conditioning, implemented in PyTorch. All training and inference are performed on a single NVIDIA RTX 4090 GPU.

The parameters are listed as Table~\ref{tab:parameters}.

\begin{figure*}[t]
   \centering
   \includegraphics[width=1.0\textwidth]{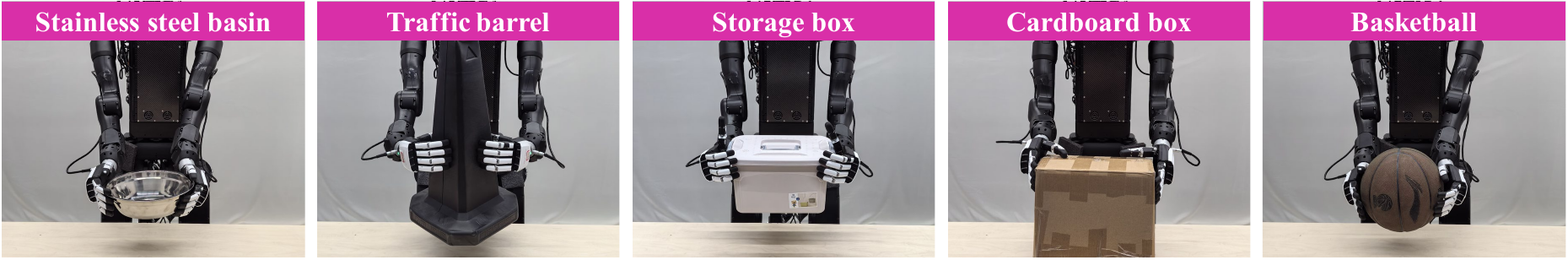}
   \caption{Example grasping poses.}
   \label{fig:each-object-grasp}
\end{figure*}

\begin{figure*}[t]
   \centering
   \includegraphics[width=1.0\textwidth]{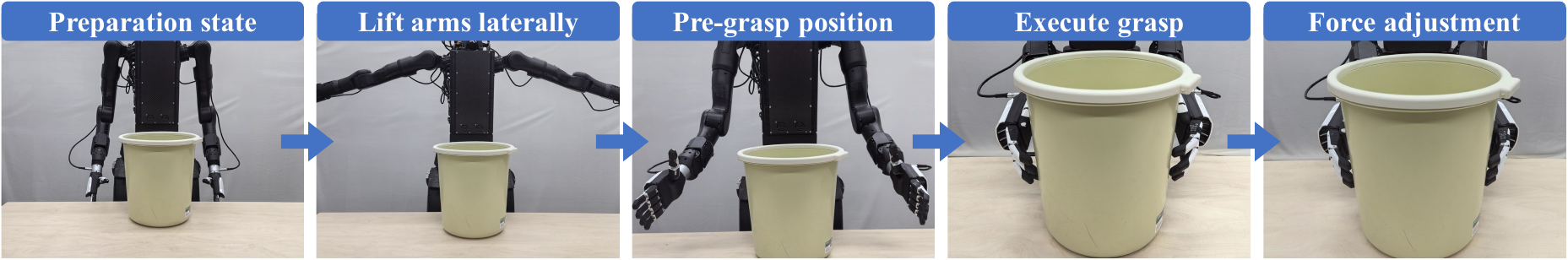}
   \caption{Demonstration of grasp execution stages.}
   \label{fig:each-stage-of-grasp}
\end{figure*}

\subsection{Evaluation Protocol and Baselines}

\begin{table*}[t]
\centering
\caption{The grasping success rates \((\%)\) comparison of our method and baseline.} 
\label{tab:rate_baseline}
\renewcommand{\arraystretch}{1.2}
\begin{tabular}{cccccccccc}
\hline
\multirow{2}{*}{\textbf{method}} & \multicolumn{8}{c}{\textbf{success rate}}                                                                                                                                            \\ \cline{2-9} 
                                 & \textbf{basin} & \textbf{barrel} & \textbf{box 1} & \textbf{box 2} & \textbf{box 3} & \textbf{bucket} & \textbf{basketball} & \multicolumn{1}{l}{\textbf{total}} \\ \hline
ours                             & \textbf{73.33}           & \textbf{53.33}           & \textbf{66.67}           & \textbf{60.00}           & \textbf{46.67}           & 60.00          & \textbf{73.33}           & \textbf{61.90}                                 \\
BimanGrasp-DDPM \cite{DBLP:journals/ral/ShaoX24_7}                          & 40.00           & 20.00           & 26.67           & 26.67           & 13.33           & 33.33           & 40.00           & 28.57                             \\ 
ViSiL-HD \cite{DBLP:journals/corr/abs-2502-20396} & 33.33           & 13.33           & 60.00           & 53.33           & 26.67           & 40.00           & 33.33           & 37.14  \\
Graspnet \cite{DBLP:conf/cvpr/FangWGL20} & \textbf{73.33}           & 13.33           & 0.00           & 0.00           & 0.00           & \textbf{66.67}           & 0.00           & 21.90  \\
DexGraspAnything \cite{DBLP:conf/cvpr/ZhongJYM25_22} & 46.67           & 26.67           & 0.00           & 0.00           & 0.00           & 36.67           & 0.00           & 15.72  \\  \hline

\end{tabular}
\end{table*}

\begin{table*}[t]
\centering
\caption{The grasping success rates \((\%)\) comparison of our method and ablation variants.} 
\label{tab:rate_ablation}
\renewcommand{\arraystretch}{1.2}
\begin{tabular}{cccccccccc}
\hline
\multirow{2}{*}{\textbf{method}} & \multicolumn{8}{c}{\textbf{success rate}}                                                                                                                                            \\ \cline{2-9} 
                                 & \textbf{basin} & \textbf{barrel} & \textbf{box 1} & \textbf{box 2} & \textbf{box 3} & \textbf{bucket} & \textbf{basketball} & \multicolumn{1}{l}{\textbf{total}} \\ \hline
ours                             & \textbf{73.33}           & \textbf{53.33}           & \textbf{66.67}           & \textbf{60.00}           & \textbf{46.67}           & \textbf{60.00}           & \textbf{73.33}           & \textbf{61.90}                                 \\
ours w/o arm squeezing           & 46.67           & 40.00           & 53.33           & 46.67           & 20.00           & 40.00           & 53.33           & 42.86                                 \\
ours w/o finger adjustment       & 53.33           & 13.33           & 46.67           & 26.67           & 13.33           & 20.00           & 66.67           & 34.29                                 \\
ours w/o motion planning         & 20.00           & 26.67           & 20.00           & 26.67           & 20.00           & 26.67           & 13.33           & 21.90                                 \\
ours w/o DDPM                    & 26.67           & 20.00           & 26.67           & 33.33           & 26.67           & 33.33           & 40.00           & 29.52                                 \\ \hline
\end{tabular}
\end{table*}

To evaluate the generalization capability to unseen objects, we adopt an object-level split of the dataset: 80\% of the objects are used for training and the remaining 20\% for testing. Each test object is placed in 5 distinct configurations on a tabletop. These configurations vary in position and orientation to reflect real-world placement variance. For every configuration, we attempt 3 independent grasps, leading to a total of 15 trials per object. All experiments are conducted under consistent lighting and background conditions.

Grasp success is defined by physical stability: a grasp is deemed successful if the robot lifts the object at least 5 cm off the table, holds it without visible slip or reorientation for 2 seconds, and completes the task within a 30-second timeout. Evaluation is binary per trial. For each test object, the success rate is averaged across 15 trials (3 attempts for each of 5 poses).

We compare our method with BimanGrasp-DDPM \cite{DBLP:journals/ral/ShaoX24_7}, which predicts hand poses in simulation. To adapt it for real-world use, we implement hand-eye calibration and inverse kinematics to convert poses into joint angles. The model is trained using the simulation dataset provided by BimanGrasp-DDPM itself, and no path planning or online adjustment is applied during execution, in order to provide a comparative demonstration of the necessity of the real-world dataset and execution pipeline proposed in this work. Additionally, we reduce each Shadow Hand to 6 controllable degrees of freedom to match our RH56DFTP platform.

We also compare our method with ViSiL-HD \cite{DBLP:journals/corr/abs-2502-20396}, a real-world approach that achieves dual-arm dexterous manipulation through reinforcement learning. This method does not achieve generalizable grasping across diverse objects; therefore, we can only use the policy it trained for grasping cuboid objects. In addition, in our experiments, the object is placed directly on the tabletop, rather than being elevated by placing a smaller box underneath, as done in their work.

Meanwhile, to demonstrate the necessity of dual-handed grasping, we compare our approach with representative single-hand grasping methods, including GraspNet \cite{DBLP:conf/cvpr/FangWGL20}, a classic 6D pose grasping algorithm based on a two-finger parallel gripper, and DexGraspAnything \cite{DBLP:conf/cvpr/ZhongJYM25_22}, a state-of-the-art single dexterous hand grasping method. Since we do not have a two-finger gripper, we emulate two-finger grasping using the multi-finger structure of the dexterous hand, where the thumb cooperates with the other fingers to achieve an equivalent parallel-grip configuration.

We conduct ablation studies by progressively removing key modules from the grasp execution pipeline. These include disabling the arm squeezing stage, the force-guided finger adjustment, and the motion planning component. Additionally, we replace the learned DDPM-based grasp synthesis with a handcrafted baseline that positions both hands symmetrically around the object’s centroid using a fully open posture. These variants allow us to isolate the effects of reactive refinement, planning, and learned coordination.

\subsection{Experimental Results and Analysis}

As shown in Table~\ref{tab:rate_baseline}, our method achieves better success rates across most test objects, indicating comparatively stable grasp execution under various shapes and placements. Figure~\ref{fig:each-object-grasp} shows the grasping poses generated by our model for several objects (without force-guided refinement), and Figure~\ref{fig:each-stage-of-grasp} illustrates the execution process of our pipeline. “Storage box” in the figure corresponds to “box 1” in the table, and “Cardboard box” corresponds to “box 2”.

The baseline method BimanGrasp-DDPM \cite{DBLP:journals/ral/ShaoX24_7} performs poorly in real-world settings for several reasons. First, it relies on an impractical assumption of access to complete object models, making it difficult to predict reasonable grasp poses from partial observations. Second, it generates only a single static grasp pose without considering the actual physical environment; as a result, issues such as pre-grasp collisions or kinematically unreachable poses often occur during execution. Moreover, the method lacks online adaptation strategies, leading to unstable contacts. In addition, cumulative errors arise from extra hand–eye calibration, inverse kinematics computation, and the mapping between different hand configurations.

\begin{figure}[t]
   \centering
   \includegraphics[width=3.3in]{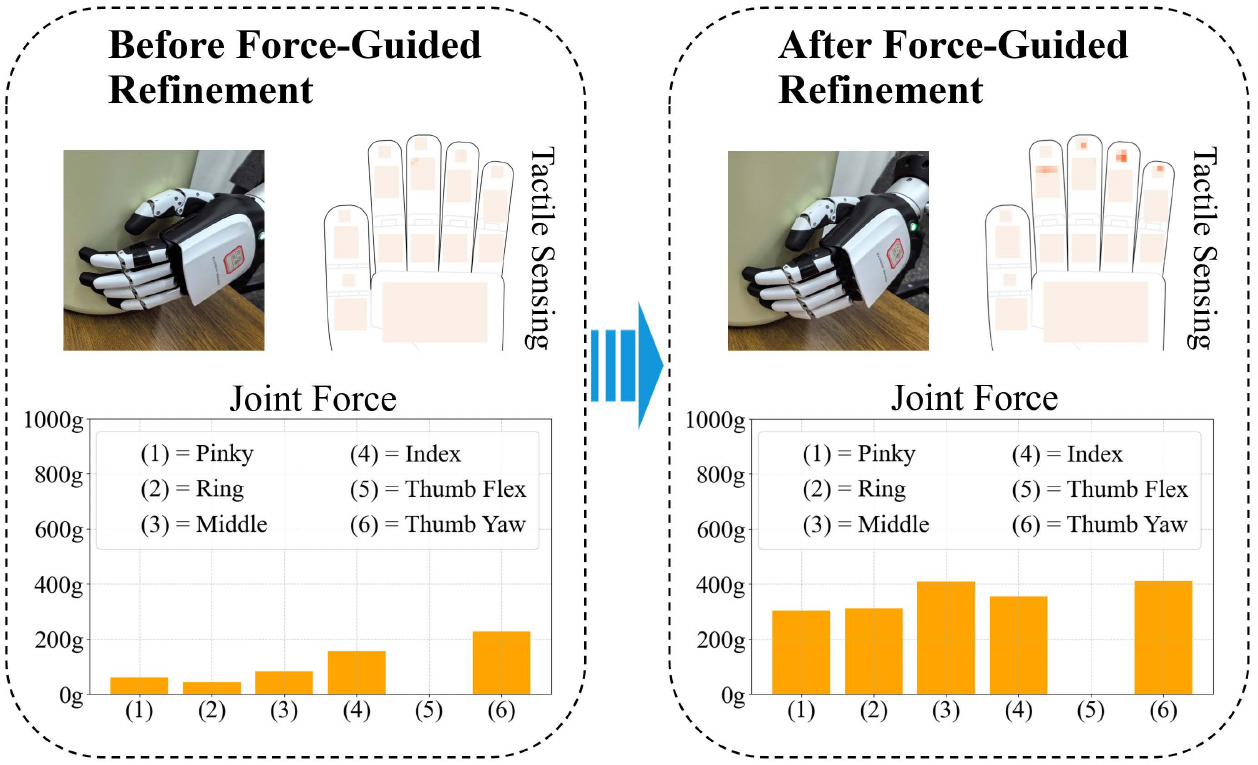}
   \caption{Comparison of dexterous hand poses, joint forces, and tactile forces before and after force-guided refinement.}
   \label{fig:Force-Visualization}
\end{figure}

The baseline method ViSiL-HD \cite{DBLP:journals/corr/abs-2502-20396} suffers from limited generalization, as it is trained exclusively on cuboid objects. Consequently, it performs poorly on non-cuboid objects. Even on box-like cuboid objects, its performance is unsatisfactory, since the learned policy partially relies on providing support from underneath the target object. However, in our tabletop experimental setup, there is no free space beneath the object that allows such bottom-side insertion.

Single-hand grasping methods GraspNet \cite{DBLP:conf/cvpr/FangWGL20} and DexGraspAnything \cite{DBLP:conf/cvpr/ZhongJYM25_22} achieve moderate success on “basin” and “bucket”, with GraspNet even outperforming our method on “bucket”. This is because their thin, vertically distributed outer edges can be pinched by a single dexterous hand. For “barrel”, although some regions can be enveloped, a single hand cannot provide sufficient force to support its weight, leading to low success rates. Large, simple-shaped objects such as “box” and “basketball” lack effective single-hand grasping points, causing all single-hand methods to fail. These results highlight the necessity of dual-arm grasping.

Removing individual modules leads to distinct failure patterns, as shown in Table~\ref{tab:rate_ablation}. Without arm squeezing or force-guided finger adjustment, the grasp may lack sufficient contact tightness or adaptability, resulting in unstable lifts. Omitting the motion planning stage causes frequent collisions with the tabletop or neighboring regions, and the arms may push or deflect the object rather than enclosing it. The handcrafted baseline, which positions hands symmetrically around the object center with open palms, fails to account for shape-specific contact and often leads to poor alignment and weak grasp formation.

Figure~\ref{fig:Force-Visualization} shows the dexterous hand poses, finger joint forces, and fingertip tactile forces before and after the force-guided refinement. It can be observed that before this adjustment, the finger joint forces are very small (not zero due to the intrinsic mechanical properties of the device and random disturbances), while the tactile sensors at the fingertips almost register no force. After the process of moving Arms Inward until Blocked and bending Fingers until Contact, not only do the finger joint forces increase significantly, but the fingertip tactile sensors also exhibit noticeable forces. This confirms the effectiveness of the force-guided refinement method, which substantially enriches and tightens the contact between the dexterous hand and the object.

\begin{figure}[t]
    \centering

    \begin{subfigure}{0.11\textwidth}
        \centering
        \includegraphics[width=\linewidth]{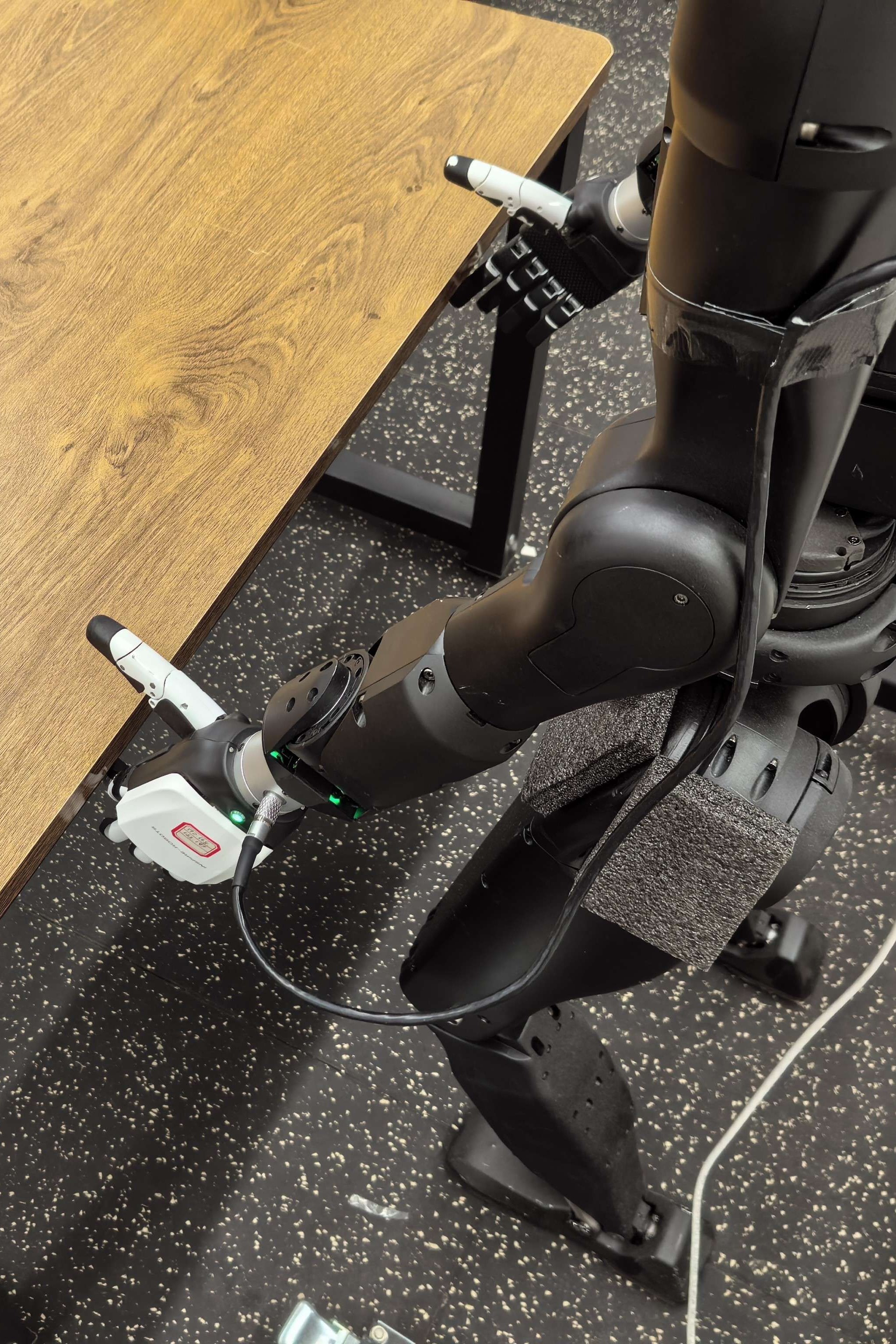}
        \caption{}
        \label{fig:a}
    \end{subfigure}
    \hfill
    \begin{subfigure}{0.11\textwidth}
        \centering
        \includegraphics[width=\linewidth]{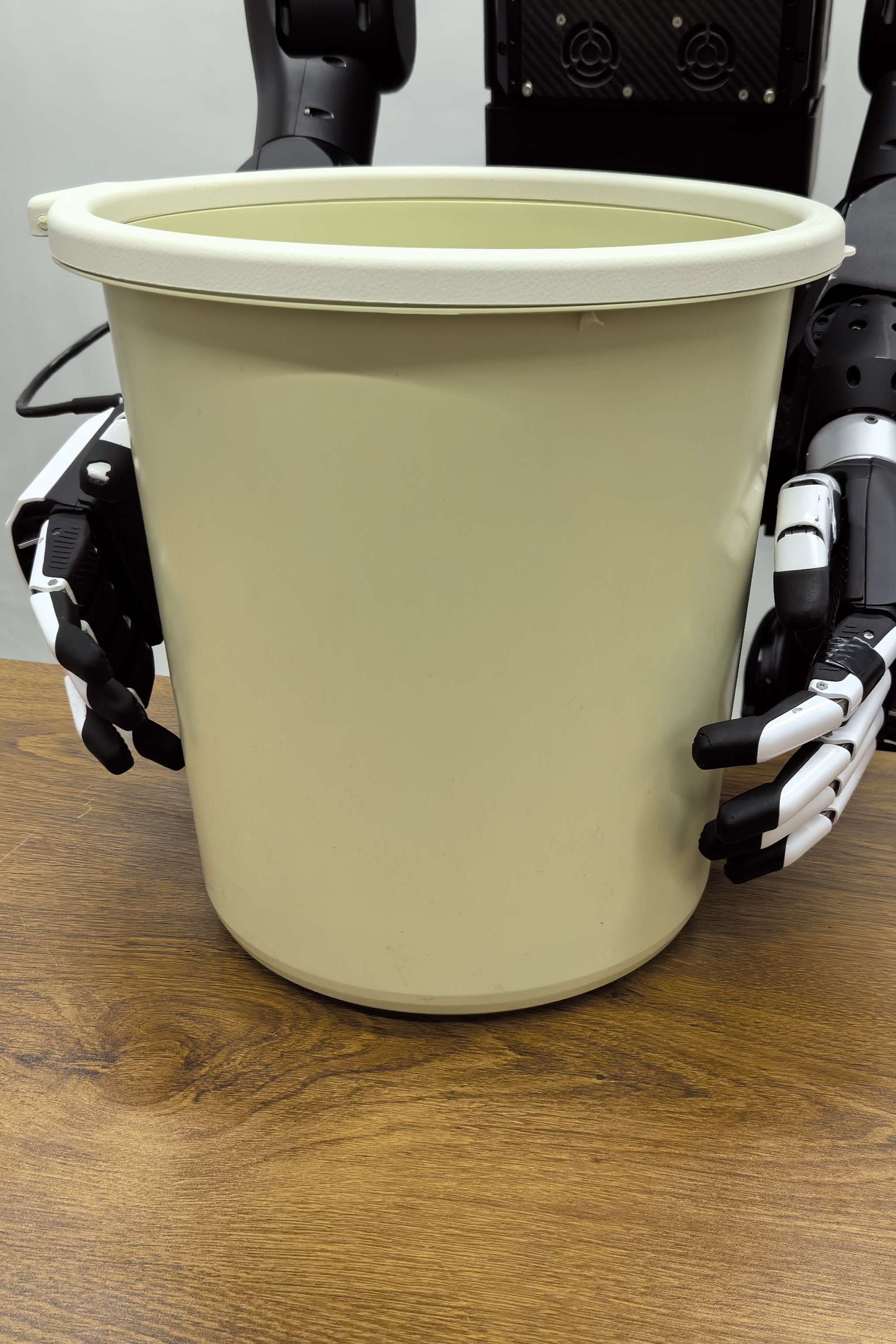}
        \caption{}
        \label{fig:b}
    \end{subfigure}
    \hfill
    \begin{subfigure}{0.11\textwidth}
        \centering
        \includegraphics[width=\linewidth]{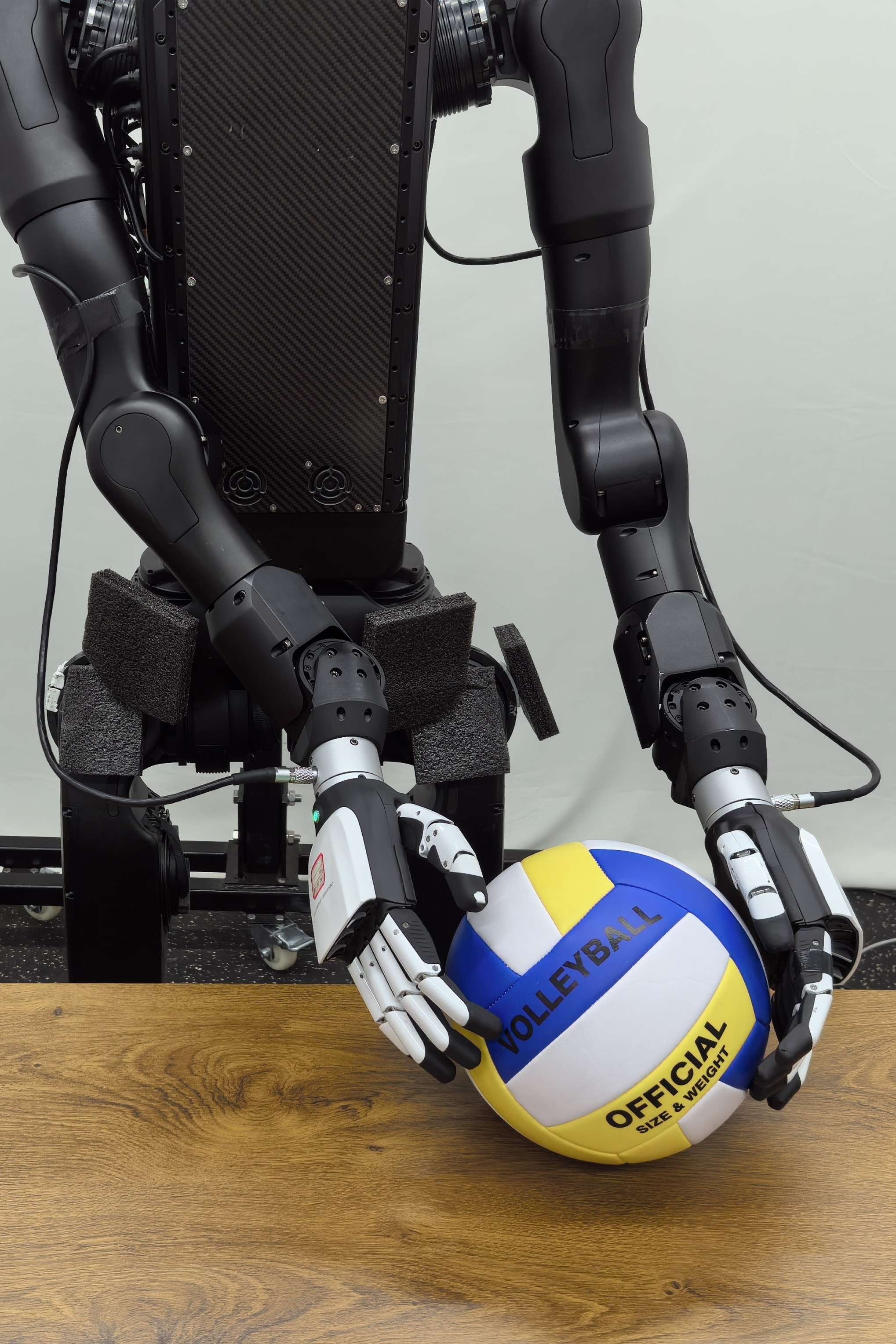}
        \caption{}
        \label{fig:c}
    \end{subfigure}
    \hfill
    \begin{subfigure}{0.11\textwidth}
        \centering
        \includegraphics[width=\linewidth]{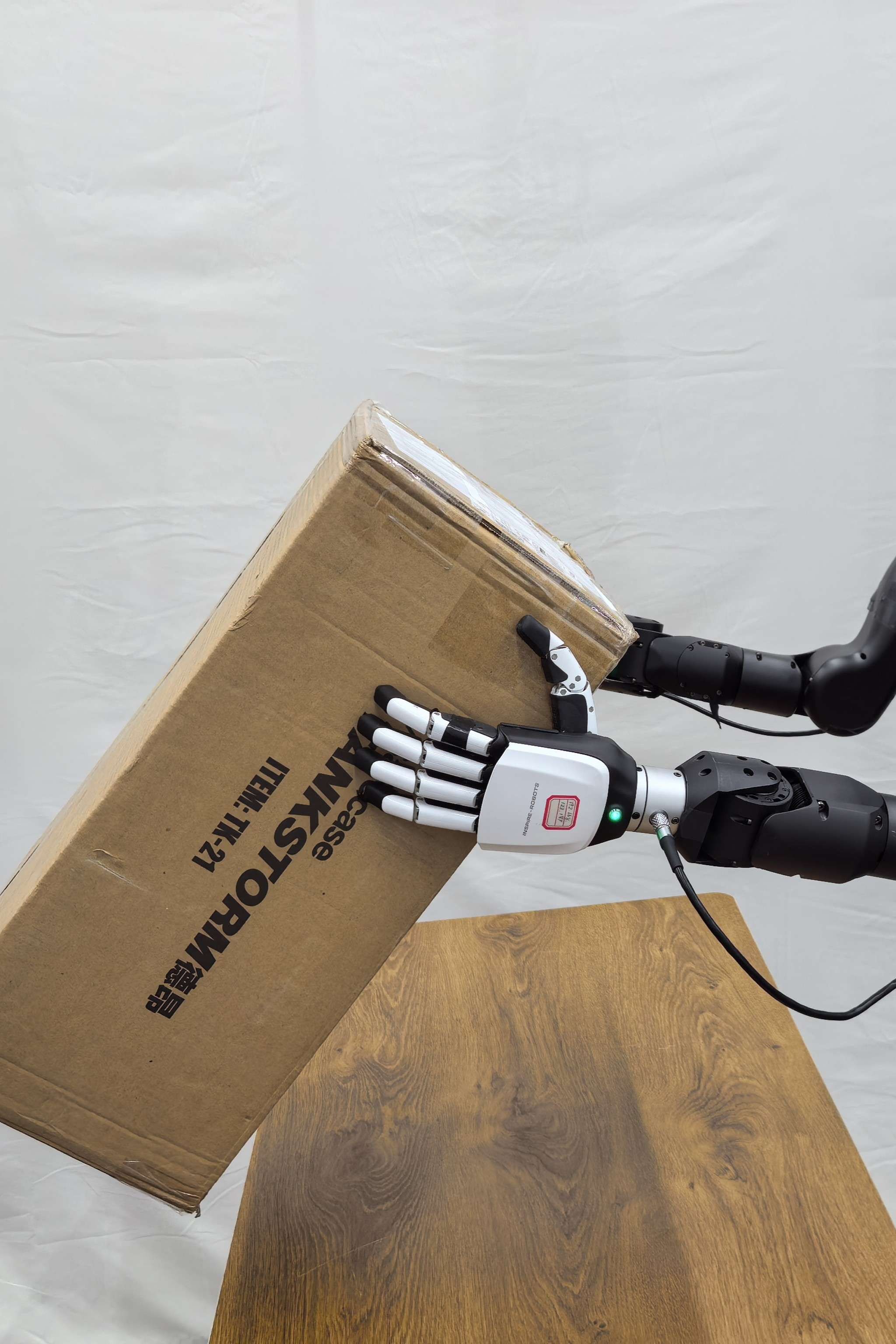}
        \caption{}
        \label{fig:d}
    \end{subfigure}

    \caption{Failure cases of grasping observed in ablation studies and challenging scenarios.}
    \label{fig:fail}
\end{figure}

Our method also exhibits occasional failures, mainly due to extreme object placements and unfavorable object size, weight, or mass distribution. Figure~\ref{fig:fail} shows several failure cases. In Figure~\ref{fig:a}, the ablation experiment without lift arms laterally is illustrated: the hands move directly from the natural hanging position to the target location, causing them to be blocked by the table. In Figure~\ref{fig:b}, the ablation experiment without force-guided refinement is shown: the hands fail to establish tight contact with the object, making it impossible to stably lift it. In Figure~\ref{fig:c}, the failure is caused by the object being placed too far to one side. Due to the obstruction of the robot’s torso, the right hand cannot reach the target grasping position, so it cannot grasp the object effectively. In Figure~\ref{fig:d}, the object is large and heavy (“box 3” in Table~\ref{tab:rate_baseline}), while the dexterous hands grasp it at one end rather than at the center, so the fingertip friction is insufficient to support the weight of the other side during lifting, leading to relative slippage of the object.

\section{Conclusions}

We present a real-world framework for bimanual dexterous grasping that leverages a teleoperated dual-arm platform to collect a multimodal dataset encompassing multimodal
 information. A diffusion-based model is trained to generate joint-level grasp poses from segmented point clouds, and a robust execution strategy is designed with motion planning and force-guided refinement to ensure successful physical interaction. The proposed pipeline shows promising performance across the tested unseen objects and configurations.

In future work, we aim to move beyond static pose prediction by developing models that generate full grasp execution trajectories. Additionally, we plan to incorporate tactile and force feedback directly into the learning process, enabling more adaptive and physically grounded grasp synthesis.

\bibliographystyle{IEEEtran}
\bibliography{reference}

\end{document}